\documentclass[runningheads]{llncs}
\usepackage[T1]{fontenc}
\usepackage{graphicx}
\usepackage{hyperref}
\usepackage{cite}
\usepackage{booktabs}
\usepackage{amsmath}
\usepackage{pgfplots}
\usepackage{makecell}
\usepackage{multirow}
\usepackage{bbm}
\usepackage{amssymb}
\usepackage{graphicx}
\usepackage{adjustbox}

\pgfplotsset{compat=1.18}

\usepackage[misc]{ifsym}
\begin{document}
\title{Harnessing Diverse Perspectives: A Multi-Agent Framework for Enhanced Error Detection in Knowledge Graphs}
\titlerunning{A Multi-Agent Framework for Enhanced Error Detection in KGs}
\author{Yu Li\inst{1}\and
Yi Huang\inst{2} \textsuperscript{(\Letter)} \and
Guilin Qi\inst{1} \textsuperscript{(\Letter)} \and
Junlan Feng\inst{2} \and
Nan Hu\inst{1} \and
Songlin Zhai\inst{1} \and \\
Haohan Xue\inst{1} \and
Yongrui Chen\inst{1} \and
Ruoyan Shen\inst{1} \and
Tongtong Wu\inst{3}}
\authorrunning{Y. Li et al.}
\institute{Southeast University, Nanjing, China\\
\email{\{yuli\_11, gqi, nanhu, songlin\_zhai, thex1ay, yongruichen, ry\_shen\}@seu.edu.cn}\\
\and
China Mobile Research Institute, China\\
\email{\{huangyi, fengjunlan\}@chinamobile.com}
\and
Monash University, Australia\\
\email{tongtong.wu@monash.edu}
}

\maketitle             
\setcounter{footnote}{0}
\begin{abstract}
Knowledge graphs are widely used in industrial applications, making error detection crucial for ensuring the reliability of downstream applications. 
Existing error detection methods often fail to effectively utilize fine-grained subgraph information and rely solely on fixed graph structures, while also lacking transparency in their decision-making processes, which results in suboptimal detection performance.
In this paper, we propose a novel Multi-Agent framework for Knowledge Graph Error Detection (MAKGED) that utilizes multiple large language models (LLMs) in a collaborative setting. 
By concatenating fine-grained, bidirectional subgraph embeddings with LLM-based query embeddings during training, our framework integrates these representations to produce four specialized agents. These agents utilize subgraph information from different dimensions to engage in multi-round discussions, thereby improving error detection accuracy and ensuring a transparent decision-making process.
Extensive experiments on FB15K and WN18RR demonstrate that MAKGED outperforms state-of-the-art methods, enhancing the accuracy and robustness of KG evaluation. 
For specific industrial scenarios, our framework can facilitate the training of specialized agents using domain-specific knowledge graphs for error detection, which highlights the potential industrial application value of our framework.
Our code and datasets are available at \url{https://github.com/kse-ElEvEn/MAKGED}.
% \footnote{This research was funded by Southeast University-China Mobile Research Institute Joint Innovation Center }.

\end{abstract}

\section{Introduction}

Knowledge graphs (KGs) \cite{kg} represent facts in the real world as triples, such as \textit{(Paris, capital\_of, France)}, facilitating the organization and scaling of information \cite{kg-info}, and have gained paramount importance in knowledge-based systems, such as retrieval-augmented generation and recommendation systems~\cite{recommend}.
However, most large-scale KGs \cite{kg}, built using rule-based methods and statistics-based methods to extract web data, often contain noisy or incorrect triples. For instance, the widely used knowledge graph NELL \cite{survey} contains around 600K incorrect triples, which account for 26\% of the set of triples in NELL. Most knowledge graph-driven tasks assume all triples are correct \cite{survey}, overlooking the impact of errors, which significantly degrades the performance of downstream tasks. This highlights the urgent need for effective KG error detection methods.

Existing methods can be broadly categorized into two types based on their evidence utilization. First, explicit evidence-based methods extract paths or subgraphs from the knowledge graph that support the truth of triples and use these explicit evidences to predict the credibility of triples~\cite{shi}. Second, embedding-based methods predict the credibility of triples by embedding entities and relations into vector space and calculating the embedded representation of paths or subgraphs~\cite{kgttm,embedding2}. Moreover, when combined with contrastive learning and pre-trained models, their performance has shown significant improvement~\cite{kg-bert,cca,sesicl}.

However, these methods have two main limitations: (1) They evaluate triples from a single, fixed perspective, either relying on static structural patterns (e.g., static graph embeddings) or a unidirectional semantic method (e.g., text embeddings). This means they assume predefined, unchanging connections between entities, without considering the dynamic context of the triples being evaluated. As a result, they struggle to adapt when the graph structure or textual representations change.
(2) They lack transparency in the evaluation process, typically providing a single confidence score without sufficient context about how decisions are made. These two limitations hinder error identification and impede model improvement. 

\begin{figure*}[t]
  \centering
  \includegraphics[width=1\textwidth]{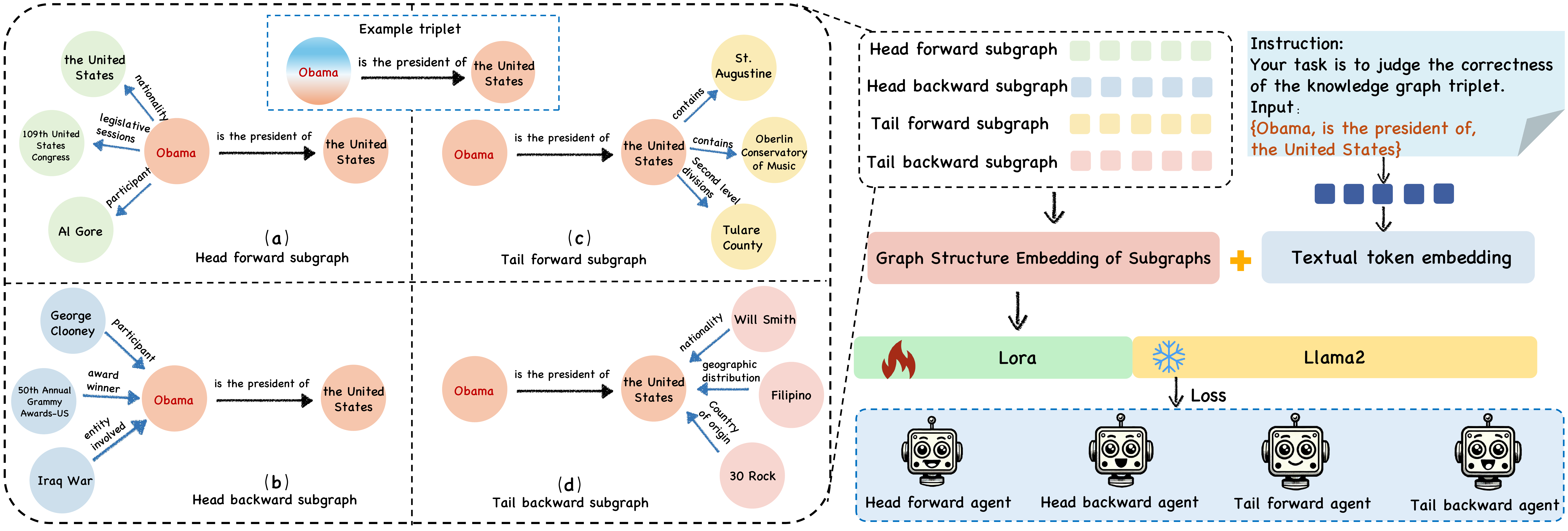}
  \caption{
  Training method for bidirectional subgraph agents in MAKGED: We first construct bidirectional subgraphs for the head and tail entities of the triple, which are represented as embedding vectors using GCN. These embeddings are then concatenated with the query embeddings of the Llama2 model, resulting in four bidirectional subgraph agents to evaluate the correctness of the triples.
  }
    \label{framework}
\end{figure*}

To address the limitations of existing methods, we introduce a multi-agent framework, as shown in Figure \ref{framework}.
For each triple in the KG, we assign two agents to the head and tail entities. The \textit{Forward Agent} collects subgraphs with the entity as the head, and the \textit{Backward Agent} collects subgraphs with it as the tail, integrating multiple perspectives for error detection.
Then, we process the collected subgraphs using a Graph Convolutional Network (GCN) for structural features and an LLM for semantic features. By concatenating the GCN and LLM embeddings, we combine structural and semantic information, leveraging the LLM's text generation capabilities for error detection.

Moreover, to address the lack of transparency during evaluation, we introduce a structured discussion and voting mechanism. The multiple agents independently evaluate each triple and then engage in multiple rounds of discussion until they reach a final decision. 
This method enhances detection accuracy and increases transparency in the decision-making process. 
At the same time, our framework can utilize domain-specific knowledge graphs to train specialized agents for industrial scenarios, improving error detection accuracy. It addresses the limitations of existing LLMs, which struggle with domain-specific knowledge, thereby delivering significant industrial value.
The main contribution are summarized as follows:

\begin{enumerate}
    \item To our knowledge, we are the first to introduce a multi-agent framework, MAKGED, for knowledge graph error detection. Agents engage in several rounds of discussion and vote to determine triple correctness, thus enhancing the clarity and accuracy of decisions through collaborative evaluation.
    \item We design four trainable agents, including the ``\textit{Forward-Agent}'' and ``\textit{Backwa-
    rd-Agent}'' for the head entity and tail entity, respectively. These agents are trained on bidirectional subgraph embeddings, which enhances the robustness and adaptability of error detection.
    \item We build datasets with simulated graph noise, and experiments demonstrate that MAKGED improves accuracy by 0.73\% on FB15K and 6.62\% on WN18RR compared to the state-of-the-art methods. 
\end{enumerate}

\section{Related Work}

\textbf{Knowledge Graph Error Detection: } 
Knowledge graph error detection~\cite{survey} includes methods based on explicit and implicit evidence.

\textbf{Explicit Evidence :} These methods evaluate the credibility of triples using direct evidence from the graph, such as paths and subgraphs. For example:
Shi et al.~\cite{shi} used graph traversal to extract meta-paths and calculate path weights.
Lin et al.~\cite{lin} applied graph pattern mining with metrics like support and confidence. However, these methods may struggle with incomplete knowledge graphs and underutilize contextual information.

\textbf{Implicit Evidence :} These methods use embedding models to map entities and relations into vector spaces for credibility evaluation. 
CKRL~\cite{CKRL} optimized triple credibility by considering local and global factors during embedding learning.
KGTtm~\cite{kgttm} combined random walks and embeddings, using MLP for scoring. Recent methods also use pre-trained language models and contrastive learning:
KG-BERT~\cite{kg-bert} enriches semantic representations with pre-trained models.
SeSICL~\cite{sesicl}, StAR~\cite{star}, and CCA~\cite{cca} align textual and structural representations via contrastive learning. These methods improve performance but rely on fixed graph structures, limiting adaptability to new or unseen structures.

\noindent\textbf{Multiple LLM-based Agent System: } 
Multi-Agent systems such as CAMEL
~\cite{camel}, have gained significant attention for simulating complex interactions among intelligent agents. These systems coordinate multiple LLM agents to tackle complex tasks, particularly in NLP evaluation.
For example, ChatEval \cite{chateval} uses a multi-agent jury to explore the impact of communication strategies on evaluating open-ended questions and traditional NLG tasks, while MATEval \cite{mateval} involves agents collaborating to evaluate story texts, enhancing the evaluation process through multi-agent interaction.

\section{Problem Statement}

\subsection{Knowledge Graph Error}

We define a knowledge graph \( \mathcal{G} \) as a set of triples \( (h, r, t) \), 
where \( h \) is the head entity, \( r \) the relation, and \( t \) the tail entity. Error detection involves determining if a given triple is correct or incorrect, with the output being a binary label.

A triple is \textbf{incorrect} if the head or tail entity does not align with the relation \cite{error}. For example, \textit{(Harvard University, is\_located\_in, New York)}. Conversely, a triple is \textbf{correct} if all components align appropriately.

\subsection{Subgraph Definitions}

To analyze the context of a triple, we define two key concepts for each entity: \textit{Out\_Neighbor Subgraph} and \textit{In\_Neighbor Subgraph}.

\noindent\textbf{\textit{Out\_Neighbor Subgraph}}: The set of triples where the entity serves as the head. For an entity \( e \), the \textit{Out\_Neighbor Subgraph} is \( \{(e, r', t') \mid (e, r', t') \in \mathcal{G}\} \), where \( r' \) is outgoing relations from \( e \), and \( t' \) is the corresponding tail entity.

\noindent\textbf{\textit{In\_Neighbor Subgraph}}: The set of triples where the entity serves as the tail. For an entity \( e \), the \textit{In\_Neighbor Subgraph} is \( \{(h', r', e) \mid (h', r', e) \in \mathcal{G}\} \), where \( h' \) is the corresponding head entity, and \( r' \) represents incoming relations to \( e \).

Based on these concepts, for a given triple \( (h, r, t) \), we define the following subgraphs for both the head \( h \) and the tail \( t \):

\noindent\textbf{(a) \textit{Head\_Forward\_Subgraph}}: The \textit{Out\_Neighbor Subgraph} of the head entity \( h \), excluding the current triple \( (h, r, t) \). Formally:
\begin{align}
   \{(h, r', t') \mid (h, r', t') \in \mathcal{G}, (r', t') \neq (r, t)\}
\end{align}
   
\noindent\textbf{(b) \textit{Head\_Backward\_Subgraph}}: The \textit{In\_Neighbor Subgraph} of the head entity \( h \), capturing all incoming relations to \( h \). Formally:
\begin{align}
   \{(h', r', h) \mid (h', r', h) \in \mathcal{G} \}
\end{align}

\noindent\textbf{(c) \textit{Tail\_Forward\_Subgraph}}: The \textit{Out\_Neighbor Subgraph} of the tail entity \( t \), capturing all outgoing relations from \( t \). Formally:
\begin{align}
   \{(t, r', t') \mid (t, r', t') \in \mathcal{G}\}
\end{align}

\noindent\textbf{(d) \textit{Tail\_Backward\_Subgraph}}: The \textit{In\_Neighbor Subgraph} of the tail entity \( t \), excluding the current triple \( (h, r, t) \). Formally:
\begin{align}
   \{(h', r', t) \mid (h', r', t) \in \mathcal{G}, (h', r') \neq (h, r)\}
\end{align}

\subsection{Agent Construction}

We construct four agents based on the above subgraphs: 
\textit{Head\_Forward\_Agent}, \textit{Head\_Backward\_Agent}, \textit{Tail\_Forward\_Agent} and \textit{Tail\_Backward\_Agent}.
Each agent analyzes the corresponding subgraph for the triple \( (h, r, t) \), enabling a multi-angle evaluation of the triple by considering both head and tail entities' forward and backward contexts.

\section{Method}

\subsection{Design of the Framework}

In our framework, we employ multiple LLM-based agents working collaboratively to detect errors in KGs. Using the structural information of the graph, we construct four bidirectional subgraph agents for both the head and tail entities. These agents analyze the contextual information of triples from different perspectives, and a final decision on the correctness of the triples is made through a voting mechanism. The detailed explanation of this process is provided below:

\textbf{Bidirectional Subgraph Agents: }
In our MAKGED framework, we design four bidirectional subgraph agents to evaluate triples in the knowledge graph. Each of these agents is responsible for analyzing triples from a specific directional perspective, including the \textit{Head\_Forward\_Agent} and \textit{Head\_Backward\_Agent} for the head entity, and the \textit{Tail\_Forward\_Agent} and \textit{Tail\_Backward\_Agent} for the tail entity, as illustrated in Figure~\ref{framework}.

First, we construct bidirectional subgraphs for both the head and tail entities of each triple. For the head entity, the \textit{Head\_Forward\_Agent} extracts the \textit{Out\_Neighbor subgraph}, where the edges represent outgoing relations from the head entity; concurrently, the \textit{Head\_Backward\_Agent} extracts the \textit{In\_Neighbor subgraph}, where the edges represent incoming relations directed toward the head entity. Similarly, for the tail entity, the \textit{Tail\_Forward\_Agent }and \textit{Tail\_Backward-
\_Agent} generate forward and backward subgraphs, representing the tail entity as either a head node or a tail node in related subgraphs.

\begin{figure*}
  \includegraphics[width=1\textwidth]{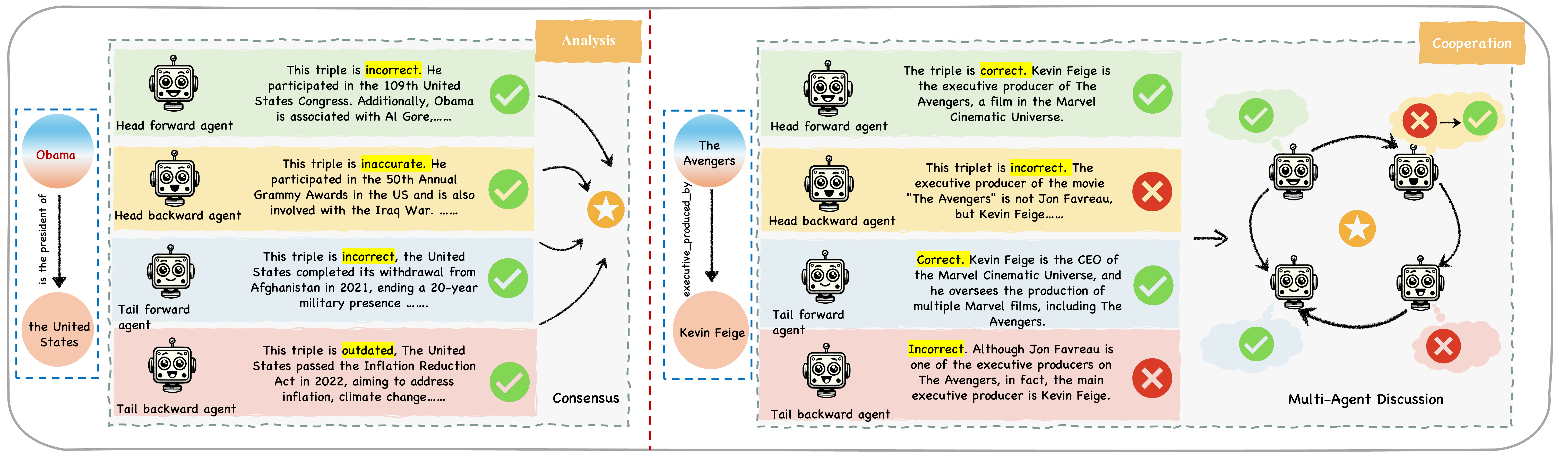}
  \caption{This figure illustrates the collaborative decision-making process using multiple agents. In the "analysis" phase, the four agents independently evaluate the triple. If no consensus is reached, they proceed to the "cooperation" phase for discussion. The final decision is made either by majority rule after three rounds of discussion, or by a summarizer in case of a  \textit{2-vs-2} tie.}
  \label{agent}
\end{figure*}

Once the subgraphs are constructed, we process them using a Graph Convolutional Network to generate the corresponding subgraph embedding vectors. Let the subgraph embeddings be denoted as \( \mathbf{z}_{G} \). These subgraph embeddings are then concatenated with the embedding vectors generated by Llama2, denoted as \( \mathbf{e}_{text} \), which provides textual information for assessing the correctness of the triples. Llama2 embeddings provide the textual representation of the triples, while the structural information from the subgraph embeddings adds complementary context. By concatenating both semantic and structural information, we create a richer, more expressive unified embedding representation:

\begin{align}
    \mathbf{e}_{concat} = [\mathbf{z}_{G} ; \mathbf{e}_{text}]
\end{align}

where \(\mathbf{z}_{G}\) denotes the graph-based embeddings generated by the GCN module. \(\mathbf{e}_{text}\) represents the semantic embeddings derived from Llama2.
Next, these concatenated embeddings are used as input to further fine-tune the Llama2 model. During fine-tuning, the model learns not only how to combine textual and structural embeddings to improve its accuracy but also how to optimize its decision-making based on the distinct features of the subgraphs in each direction. The input sequence to the model is defined as:
\begin{align}
    S_{it} = I_{it} \oplus \mathbf{e}_{concat} \oplus A_{it},
\end{align}
where \( I_{it} \) is the instruction prompt, \( \mathbf{e}_{concat} \) is the concatenated embedding from both the GCN and Llama2, and \( A_{it} \) is the predicted answer during training. 
% We compute the loss by comparing the model's output with the ground truth labels. 
The training objective is to minimize the following loss function:

\begin{align}
    \mathcal{L}_{it} = - \frac{1}{|S_{it}|} \sum_{i=1}^{|S_{it}|} \log P_{\mathcal{M}}(s_i \mid s_{<i}, \mathbf{e}_{concat}), 
\end{align}
where \( |S_{it}| \) represents the length of the input sequence \( S_{it} \), and \( s_i \) is the token at position \( i \) in the input sequence. \( P_{\mathcal{M}}(s_i \mid s_{<i}, \mathbf{e}_{concat}) \) is the probability distribution predicted by the model for token \( s_i \), conditioned on all previous tokens and the concatenated embedding.
During the training process, we simultaneously train the Llama2 to evaluate the correctness of triples in scenarios where reasoning is provided. In this part, subgraph information is incorporated into the input, allowing the model to fine-tune its ability to discuss the correctness of triples based on reasoning.

As a result, we train four specialized agents, each tailored to specific directional tasks for either the head or tail entities (forward or backward). This method allows us to comprehensively evaluate the correctness of triples from multiple directions, significantly enhancing the performance and accuracy.

\textbf{Agent Decision: }
The agents trained in the previous process are used for the KG error detection task on the test set. This process is divided into two phases: the analysis phase and the cooperation phase. In the analysis phase, the four agents (\textit{Head\_Forward\_Agent} and \textit{Head\_Backward\_Agent} for the head entity, and \textit{Tail\_Forward\_Agent} and \textit{Tail\_Backward\_Agent} for the tail entity) evaluate the correctness of a target triple independently, making full use of the corresponding subgraph information they learned during training and minimizing mutual interference.

After collecting the results, a consistency check is performed. If all agents agree on the correctness of the triple (i.e., consensus), it is classified as correct or incorrect. If there is disagreement, the process moves to the cooperation phase.

In the cooperation phase, the four agents engage in a collective discussion, exchanging their viewpoints and background knowledge to resolve disagreements regarding the triple. This discussion process iterates for up to three rounds, stopping early if consensus is reached within these rounds. After each round of discussion, the agents update their judgments. At the end of the discussion, a ``majority rule'' strategy is employed to determine the final decision. If a \textit{2-vs-2} tie still occurs after the three rounds, 
the final decision is made by a summarizer agent, which receives the full context of the three discussion rounds as a structured prompt. This prompt includes key arguments, evidence, and conclusions from all agents, enabling the summarizer to make an informed judgment that reflects the collective reasoning of the agents.
On average, in our experience, agents reached a consensus within 1.8 rounds of discussion. In about 12\% of cases, a \textit{2-vs-2} tie occurred, which was resolved by the summarizer agent.
The entire agent discussion process is illustrated in Figure \ref{agent}.

\section{Experiments}

\begin{table*}[t]
\begin{center}{
    \caption{Results on FB15K and WN18RR, comparing Accuracy, F1-Score, Precision, and Recall. We compared embedding-based methods, PLM-based methods, contrastive learning-based methods, and LLM-based methods.}
    \label{main_result}
}

    \scalebox{0.9}{
        \begin{tabular}{lcccccccc}
 
        \toprule

        \multirow{2}{*}{Models} & \multicolumn{4}{c}{FB15K} & \multicolumn{4}{c}{WN18RR} \\
        \cmidrule(lr){2-5} \cmidrule(lr){6-9}
        & Accuracy & F1-Score & Precision & Recall & Accuracy & F1-Score & Precision & Recall \\
 
        \midrule
        \multicolumn{9}{l}{\textit{Embedding-Based Methods}} \\

        TransE & 0.6373 & 0.6312 & 0.6410 & 0.6531 & 0.3813 & 0.2927 & 0.6255 & 0.5083 \\
        DistMult & 0.5938 & 0.5132 & 0.5261 & 0.5204 & 0.6401 & 0.5157 & 0.5965 & 0.5449 \\
        ComplEx & 0.6268 & 0.4781 & 0.5413 & 0.5172 & 0.6414 & 0.4450 & 0.6464 & 0.5217 \\
        CAGED & 0.6091 & 0.4574 & 0.5028 & 0.4552 & 0.6544 & 0.5064 & 0.5532 & 0.5013 \\
        KGTtm & 0.6828 & 0.4078 & 0.6172 & 0.3045 & 0.6911 & 0.4487 & 0.6589 & 0.3402 \\

        \midrule
        \multicolumn{9}{l}{\textit{PLM-based Methods}} \\

        KG-BERT & 0.7675 & 0.6280 & 0.7371 & 0.5470 & 0.8162 & 0.7222 & 0.8177 & 0.6468 \\
        StAR & 0.7350 & 0.6017 & 0.6900 & 0.5420 & 0.7012 & 0.6100 & 0.6572 & 0.5645 \\ 
        CSProm-KG & 0.7078 & 0.5509 & 0.6139 & 0.4997 & 0.7116 & 0.6025 & 0.6138 & 0.4997 \\
        
        \midrule
        \multicolumn{9}{l}{\textit{Contrastive Learning-based Methods}} \\

        SeSICL & 0.5950 & 0.4600 & 0.5513 & 0.5172 & 0.5050 & 0.4073 & 0.4421 & 0.5711 \\
        CCA & 0.7456 & 0.6810 & 0.7123 & 0.6537 & 0.7621 & 0.7134 & 0.7568 & 0.6912 \\ 

        \midrule
        \multicolumn{9}{l}{\textit{LLM-based Methods}} \\

        Llama2 & 0.7420 & 0.6010 & 0.7250 & 0.6851 & 0.7100 & 0.6271 & 0.7021 & 0.6344 \\  
        GPT-3.5 & 0.7445 & 0.6117 & 0.7185 & 0.6555 & 0.7603 & 0.7496 & 0.7120 & 0.6260 \\ 
        Llama3 & 0.7558 & 0.6264 & 0.7357 & 0.7148 & 0.7654 & 0.7522 & 0.7185 & 0.6327 \\
        
        \midrule
        \multicolumn{9}{l}{\textit{Our Methods}} \\

        MAKGED & \textbf{0.7748} & \textbf{0.7367} & \textbf{0.7686} & \textbf{0.7252} & \textbf{0.8283} & \textbf{0.7909} & \textbf{0.8832} & \textbf{0.7704} \\

        \bottomrule

        \end{tabular}
    }
    % \caption{Results on FB15K and WN18RR, comparing Accuracy, F1-Score, Precision, and Recall. We compared embedding-based methods, PLM-based methods, contrastive learning-based methods, and LLM-based methods.}
    % \label{main_result}
\end{center}
\end{table*}

To validate the effectiveness of our proposed MAKGED framework, we conducted comprehensive experiments on two representative knowledge graph datasets in this section, as well as in industrial scenarios such as China Mobile. Specifically, we aim to answer the following research questions through  experiments:

\noindent\textbf{RQ1}: How does MAKGED perform compared to state-of-the-art KG error detection methods?

\noindent\textbf{RQ2}: How does each component of the MAKGED framework contribute to its overall performance?

\noindent\textbf{RQ3}: Can the MAKGED framework successfully detect specific errors in knowledge graphs, especially in industrial applications such as those at China Mobile?

\subsection{Experimental Settings}

\noindent\textbf{Datasets: } 
We use two real-world knowledge graph datasets: FB15K \cite{fb15k} and WN18RR~\cite{wn18rr}. We chose these two datasets because they are highly representative in the field of knowledge graph error detection, encompassing most typical scenarios and possible graph structural representations found in knowledge graph data. In each dataset, we simulate realistic errors by replacing entities and relations with similar ones selected based on cosine similarity within the dataset, resulting in approximately 30\% of the data being erroneous. We split each dataset into training, validation, and test sets with a ratio of 8:1:1. The fine-tuning process used only the training set, while the test set was reserved for final evaluations. 
FB15K is derived from Freebase and contains a rich set of entities and relations, while WN18RR is a subset of WordNet with corrected inverse relations, increasing the complexity. 
Additionally, we conducted experiments on a knowledge graph dataset from China Mobile's business scenarios, achieving the best results compared to other methods.

\noindent\textbf{Baselines: } We compare MAKGED against various baseline methods, including traditional knowledge graph embedding models such as TransE \cite{transe}, DistMult \cite{distmult}, and ComplEx \cite{complex}, which learn triple embeddings to compute confidence scores. Additionally, we compared recent embedding-based KG error detection methods including CAGED \cite{CAGED} and KGTtm \cite{kgttm}. We also compared KG-BERT \cite{kg-bert} and CSProm-KG \cite{csprom} models that combine pre-trained language models for error detection, as well as models that use text structure and graph structure for comparative learning, such as StAR \cite{star}, SeSICL \cite{sesicl}, and CCA \cite{cca} models. All comparison experiments are conducted under the same experimental settings.

\noindent\textbf{Implementation Details: }
We use the Llama2 for fine-tuning and employ LoRA \cite{lora} for instruction tuning. Experiments are conducted on V100 GPU servers. 
\textbf{GCN Component:} A three-layer GCN with 128 hidden dimensions and 64-dimensional embeddings is trained separately using the Adam optimizer (learning rate: 0.001), a batch size of 64, for 100 epochs. The resulting GCN embeddings are concatenated with Llama2's text embeddings to create a unified representation for each agent.
\textbf{Fine-Tuning Llama2:} During fine-tuning, the combined embeddings (GCN + Llama2) serve as model inputs. We utilize mixed precision training and gradient checkpointing to accelerate training and reduce memory usage.
We use Accuracy, F1-Score, Precision, and Recall as evaluation metrics. These metrics use macro averaging for both classes.
Each round of agent discussion took an average of 2.3 seconds. This demonstrates the framework's practical feasibility for industrial applications.

\begin{table*}[t]
\begin{center}{
    \caption{The ablation study on FB15K and WN18RR evaluates Accuracy, F1-Score, Precision, and Recall. Results highlight the importance of bidirectional subgraph training and multi-agent discussions. }
    \label{ablation}
}

    \scalebox{0.83}{
        \begin{tabular}{lcccccccc}
 
        \toprule

        \multirow{2}{*}{Models} & \multicolumn{4}{c}{FB15K} & \multicolumn{4}{c}{WN18RR} \\
        \cmidrule(lr){2-5} \cmidrule(lr){6-9}
        & Accuracy & F1-Score & Precision & Recall & Accuracy & F1-Score & Precision & Recall \\
 
        \midrule
        \multicolumn{9}{l}{\textit{Only use a specific sub-graph for discussion}} \\

        MAKGED \scriptsize{(Head\_as\_Head)} & 0.6920 & 0.6129 & 0.6496 & 0.6098 & 0.6940 & 0.5467 & 0.7826 & 0.5799 \\
        MAKGED \scriptsize{(Head\_as\_Tail)} & 0.7220 & 0.6283 & 0.7115 & 0.6254 & 0.6920 & 0.5422 & 0.7802 & 0.5771 \\
        MAKGED \scriptsize{(Tail\_as\_Head)} & 0.7100 & 0.6144 & 0.6875 & 0.6135 & 0.7000 & 0.5571 & 0.7999 & 0.5870 \\
        MAKGED \scriptsize{(Tail\_as\_Tail)} & 0.7080 & 0.6147 & 0.6820 & 0.6133 & 0.6940 & 0.5498 & 0.7726 & 0.5811 \\

        \midrule
        \multicolumn{9}{l}{\textit{Discuss without combining sub-graph information}} \\

        Llama2 & 0.7020 & 0.6432 & 0.6627 & 0.6373 & 0.6800 & 0.5368 & 0.7064 & 0.5689 \\
        
        \midrule
        \multicolumn{9}{l}{\textit{Make direct judgments without discussion}} \\

        MAKGED \scriptsize{(Analysis)} & 0.7300 & 0.6916 & 0.7410 & 0.6750 & 0.7100 & 0.5733 & 0.7967 & 0.5955 \\
        
        \midrule
        \multicolumn{9}{l}{\textit{Use the complete framework structure}} \\

        MAKGED & \textbf{0.7748} & \textbf{0.7367} & \textbf{0.7686} & \textbf{0.7252} & \textbf{0.8283} & \textbf{0.7909} & \textbf{0.8832} & \textbf{0.7704} \\

        \bottomrule

        \end{tabular}
    }
    % \caption{The ablation study on FB15K and WN18RR evaluates Accuracy, F1-Score, Precision, and Recall. Results highlight the importance of bidirectional subgraph training and multi-agent discussions. }
    % \label{ablation}
\end{center}
\end{table*}

\subsection{Effectiveness Analysis}
\noindent\textbf{Experiment Setup:}
To study RQ1, we conducted comprehensive experiments on two KG datasets, comparing it with the previously mentioned baseline models across four key metrics. The experimental results are presented in Table \ref{main_result}.

\noindent\textbf{Comparison to Embedding-based Methods:} MAKGED combines subgraph structural information with semantic insights from LLMs, leading to a 10-20\% improvement in accuracy and a significant increase in the F1-Score.

\noindent\textbf{Comparison to PLM-based Methods: } 
MAKGED further demonstrates its strong error detection capabilities. Although models like KG-BERT show advantages in semantic understanding, their utilization of knowledge graph structure is limited. By integrating subgraph embeddings with LLMs embeddings, MAKGED improves recall by around 10\% on the WN18RR compared to KG-BERT, demonstrating better coverage and precision in detecting complex errors.

\noindent\textbf{Comparison to Contrastive Learning Methods:} MAKGED shows stronger performance by leveraging multi-agent collaboration and integrating multiple perspectives. On the FB15K dataset, MAKGED achieved an F1-Score improvement of approximately 5-8\%. 

\noindent\textbf{Comparison to LLM-based Methods:} 
While LLM-based methods, such as Llama2 and GPT-3.5, perform well in knowledge graph tasks, their limited understanding of graph structures reduces their effectiveness in error detection. MAKGED overcomes this limitation by combining the semantic power of LLMs with the structural insights of GCN, allowing for more accurate error detection. Notably, its performance rivals that of GPT-3.5 and Llama3. 
Overall, MAKGED consistently surpasses state-of-the-art models in accuracy, F1-Score, and recall, demonstrating outstanding error detection capabilities.

\begin{figure*}[t]
\centering
  \includegraphics[width=0.98\textwidth]{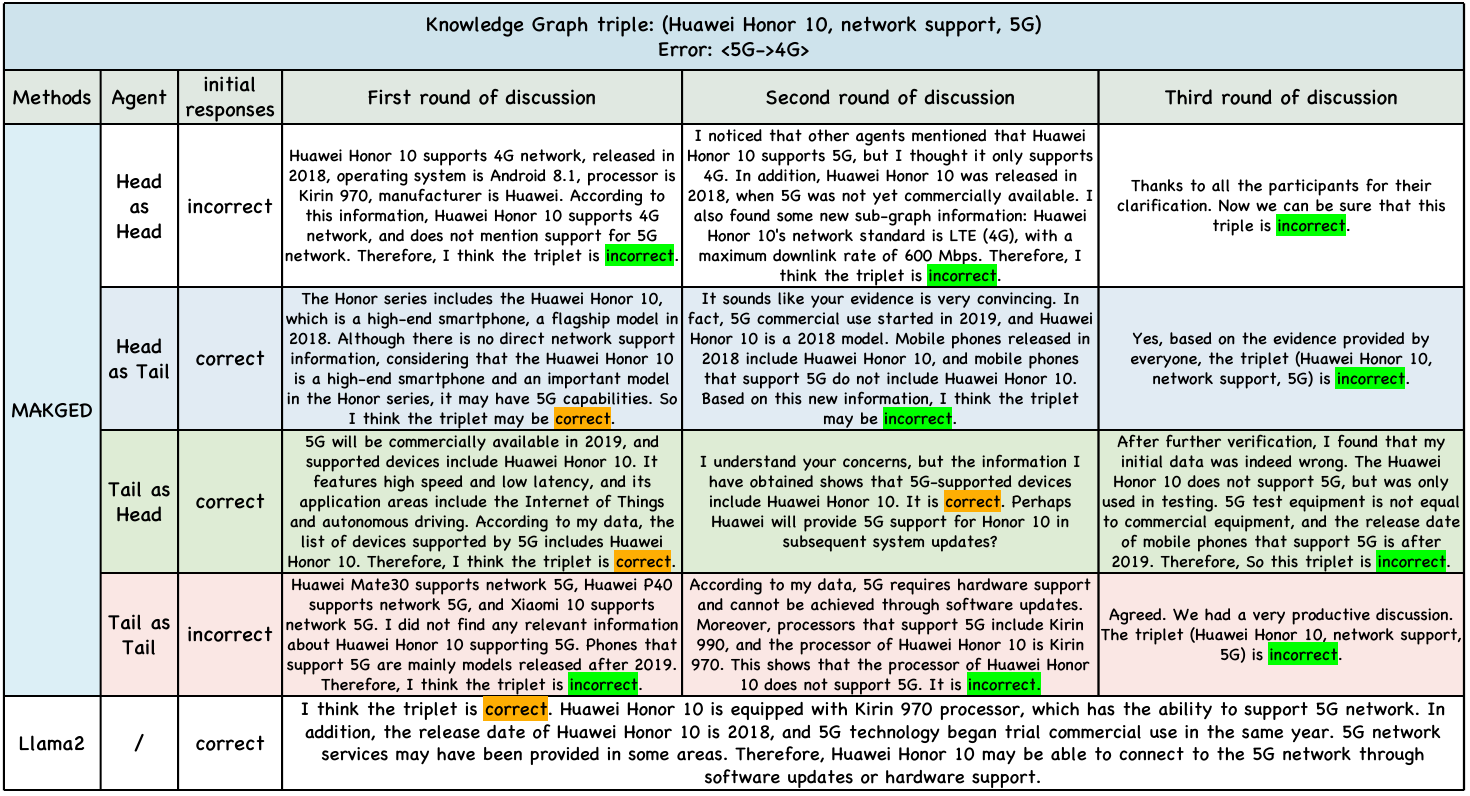}
  \caption{The figure shows an example of using our framework on the China Mobile business scenario knowledge graph. We compare the evaluation results of the original Llama2 model without subgraph fine-tuning, a method without multi-agent discussion, and our complete MAKGED framework. 
  % This example clearly demonstrates the superiority of our framework.
  }
  \label{case}
\end{figure*}

\subsection{Ablation Study}

\noindent\textbf{Experiment Setup:} 
To address RQ2, we conducted an ablation study evaluating several model variants under the same experimental setup as the full framework. The \textit{\textbf{Head\_as\_Head}} variant refers to the results where all four agents are trained using the \textit{\textbf{Head\_Forward Subgraph }}constructed from the triple’s head entity, with other ablation versions following a similar naming convention. The Analysis variant performs only initial judgments without multi-agent discussions, while the Llama2 baseline excludes subgraph information during fine-tuning, relying solely on Llama2's original outputs.

\noindent\textbf{Full Framework Outperforms Variants: } As shown in Table \ref{ablation}, when only specific subgraphs are used or subgraph information is excluded, performance drops notably compared to the complete MAKGED, especially in F1-Score and Recall, indicating that a single perspective or lack of structural information limits error detection. While using multiple agents without discussion offers slight improvements, it still lags behind the full framework.
The complete framework achieves the best results on two datasets, maximizing four metrics.

\subsection{Case Study}
\noindent\textbf{Running Example: } To study RQ3, we select an incorrect triple \textit{(Huawei Honor 10, network support, 5G)} from the industrial KG of China Mobile to demonstrate the effectiveness of our framework in industrial applications. Fig \ref{case} shows the evaluation and discussion paths for this triple under our framework. 

\noindent\textbf{Subgraph-Aided Error Correction: }
The framework effectively uses subgraph information for in-depth analysis, enabling agents to correct initial errors and reach the correct conclusion. In contrast, using the original Llama2 model without subgraph fine-tuning leads to a lower accuracy performance.

\noindent\textbf{Value of Multi-Agent Collaboration: }
If we had relied only on the initial model output, the result would have been ``correct'', conflicting with the ground truth. However, after three rounds of discussion, the agents reached the correct conclusion, demonstrating the effectiveness of multi-agent collaboration.

\section{Conclusion}

In this paper, we propose MAKGED, a novel framework for knowledge graph error detection. 
By combining subgraph embeddings from a GCN with LLM embeddings, we train four agents to evaluate triples through multi-agent discussions, enabling multi-perspective analysis. Experiments demonstrate that MAK-
GED significantly outperforms traditional and LLM-based methods, improving accuracy, F1-Score, precision, and recall across two datasets. Moreover, our framework has also shown excellent performance in industrial scenarios, validating the industrial application value of our method.

\section{Acknowledgments}
This work is partially supported by National Nature Science Foundation of China under No. U21A20488, and is funded by Southeast University-China Mobile Research Institute Joint Innovation Center. We thank the Big Data Computing Center of Southeast University for providing the facility support on the numerical calculations in this paper.

%
% ---- Bibliography ----
%
% BibTeX users should specify bibliography style 'splncs04'.
% References will then be sorted and formatted in the correct style.
%
\bibliographystyle{splncs04}
\bibliography{reference}

\end{document}